\documentclass[runningheads]{llncs}

\usepackage{eccv}

\usepackage{eccvabbrv}

\usepackage{graphicx}
\usepackage{booktabs}

\usepackage[accsupp]{axessibility}  

\usepackage{hyperref}

\usepackage{orcidlink}

\usepackage{algorithm}
\usepackage{algpseudocode}
\usepackage{xspace}
\usepackage{xcolor}
\usepackage{multirow}

\newcommand{\reid}{Re-ID\xspace}
\newcommand{\ice}{Facility-ReID\xspace}

\newcommand{\ours}{MICRO-TRACK\xspace}

\begin{document}

\title{Multi-Camera Industrial Open-Set Person Re-Identification and Tracking} 


\author{Federico Cunico\inst{1}\orcidlink{0000-0001-9619-9656} \and
Marco Cristani\inst{1}\orcidlink{0000-0002-0523-6042}}

\authorrunning{F.~Cunico, M.~Cristani}

\institute{University of Verona, 
Department of Engineering DIMI, Italy
\email{name.surname@univr.it}}

\maketitle

\begin{abstract}
 
In recent years, the development of deep learning approaches for the task of person re-identification led to impressive results.
However, this comes with a limitation for industrial and practical real-world applications.
Firstly, most of the existing works operate on closed-world scenarios, in which the people to re-identify (probes) are compared to a closed-set (gallery). Real-world scenarios often are open-set problems in which the gallery is not known a priori, but the number of open-set approaches in the literature is significantly lower. Secondly, challenges such as multi-camera setups, occlusions, real-time requirements, \etc, further constrain the applicability of off-the-shelf methods.
This work presents \ours, a \textbf{M}odular \textbf{I}ndustrial multi-\textbf{C}amera \textbf{R}e$-$identification and \textbf{O}pen-set \textbf{Track}ing system that is real-time, scalable, and easy to integrate into existing industrial surveillance scenarios. Furthermore, we release a novel \reid and tracking dataset acquired in an industrial manufacturing facility, dubbed \ice, consisting of 18-minute videos captured by 8 surveillance cameras. 

\end{abstract}

\section{Introduction}
\label{sec:intro}

Person re-identification (\reid) is a classic yet challenging task in computer vision that involves identifying the same individual across images captured by various cameras or at different times~\cite{ye2021deep}. 
Re-ID is crucial in video surveillance and security systems, where tracking individuals across different areas or locations using multiple video cameras is often necessary. For instance, it can be utilized in malls, airports, fairs, or other public places with large crowds and limited cameras \cite{becattini2022imall}.

Closed-set person re-identification is the conventional approach to this problem. In this method, the system operates with a predetermined, finite set of known identities (gallery). When the system detects a new individual (probe), it extracts features from their appearance and compares them to the gallery to identify the most similar known identity.

Open-set, on the other side, presents a more complex challenge as it deals with scenarios where the probe individual may not belong to any of the pre-registered identities in the gallery. In this kind of \reid, the system must not only match the probe to known identities but also determine when a probe does not correspond to any known individual, thus effectively managing the possibility of encountering unseen identities. This requires the development of robust algorithms capable of distinguishing between known and unknown identities while minimizing false rejections and acceptances. 
Open-set person re-identification is crucial for real-world applications such as surveillance and security, where the likelihood of encountering individuals not present in the system's gallery is high. 

\begin{figure}[t]
    \centering
    \includegraphics[width=\linewidth]{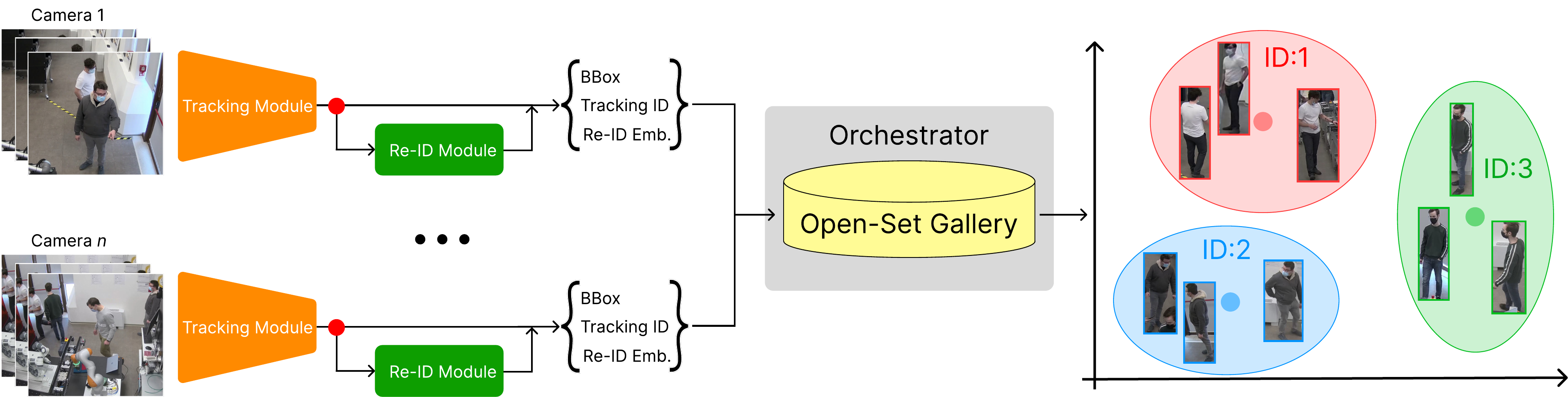}
    \caption{A schema of the multi-camera implementation of \ours. A single-camera tracking module extracts the bounding boxes and tracking ids for each camera of the environment.
    The tracking system considers the occlusions and variability in appearance through time, granting temporal consistency to the re-identification. The decision module (red dot) decides whether the person is qualitatively good enough to extract the \reid embeddings. A global Orchestrator system collects all the tracking IDs and \reid embeddings, deciding if the person has already been seen by the system or is a new entry. The Orchestrator maintains the open-set gallery.}
    \label{fig:system}
\end{figure}

While closed-set \reid is a simpler problem \wrt open-set \reid, and potentially more accurate within specific scopes, has some limitations, particularly when considering the possibility of adding new identities, which may happen in real-world industrial scenarios~\cite{leng2019survey}, such as industrial settings where the workforce may change frequently or in crowded environments.

For this reason, an open-set approach may be more interesting in industrial applications. Being more adaptable, the open-set approach can handle unknown individuals, making it better suited for industrial settings' variability and dynamic nature.

Furthermore, the system may not be scalable: as the number of individuals increases, the training set needs to be adjusted, and the need for re-training may be cumbersome and resource-intensive.
Environmental variability also plays a role. Industrial environments are typically dynamic, with varying lighting conditions, background clutter, and occlusions, which can degrade the performance of \reid systems~\cite{lei2022reducing}.

In this work, we propose a practical approach to implement a robust open-set \reid method deployable on existing industrial surveillance systems with minimum integration efforts. 
Our \ours (\textbf{M}odular \textbf{I}ndustrial multi-\textbf{C}amera \textbf{R}e-identification and \textbf{O}pen-set \textbf{Track}ing) system is designed to be modular, real-time, and robust to occlusions, even in complex dynamic environments with multiple 
cameras.
Our approach considers the multi-camera open-set \reid as a problem composed of two aspects: tracking through time and \reid using spatial appearance (see also \Cref{fig:system}). 
We leverage the robustness of people detectors to occlusion and provide people's trajectory, identified by pseudo-ids, through a reliable temporal tracking system.
Then, under conditions of good quality images, the appearance-based method for \reid provides a unique re-identification ID. 

The tracking module prevents the misclassification of identities and maintains the identity assigned by the \reid module.

This approach has been successfully tested and installed in a real-world industrial facility, where a multi-camera surveillance system has been implemented for safety reasons. The \reid method has been trained on large benchmarks and tested on a novel open-set dataset acquired in a manufacturing industrial facility. This dataset consists of 18-minute videos with 5 identities moving inside the facility with 8 cameras, for a total of $\sim210$k frames. The dataset includes several challenges for the re-identification task, such as heavy occlusions (with the environment and between people), skewed views, changes in ID appearances (same person, different clothes), and long-range points of view.

To summarize, our contributions are the following:
\begin{itemize}
    \item We propose a way to implement a robust and scalable open-set person re-identification system in industrial environments;
    \item We provide a challenging novel dataset captured in a real manufacturing facility and use it to extensively evaluate the multi-camera person \reid performance of our system.
\end{itemize}

The following sections are structured as follows: \Cref{sec:related} gives an overview of the state-of-the-art regarding the task of open-set person re-identification. \Cref{sec:method} explains how our \ours system, works. \Cref{sec:exps} present the experiments we performed to evaluate our system, both in the traditional closed-set benchmarks and in the open-set scenario, also presenting the results on the novel \ice dataset with relative discussion; in \Cref{sec:discussion} we discuss the results. \Cref{sec:limitations} highlights possible limitations of the system that may be useful in the case of industrial adoption of \ours. Finally, \Cref{sec:conclusions} presents the conclusions of this work.

\section{Related works}\label{sec:related}

\subsection{Person Re-Identification}

Recent advancements in deep learning have significantly improved person re-identification systems, yet most existing methods are tailored for closed-world scenarios. These methods compare detected people (also called query probes) against a closed-set gallery, limiting their applicability in open-set and real-world industrial environments. 
Nonetheless, several attempts were made to perform \reid in the open-set scenarios.

In the open-set person \reid scenario, \cite{vidanapathirana2017open} implemented a method using an open-set gallery, but their method depends completely on the reliability of their \reid module. In contrast, our approach applies \reid only when the images of individuals are clean, using a tracking system to minimize feature representation noise. This distinction is crucial for maintaining high accuracy in diverse and dynamic industrial settings.
Another related work~\cite{casao2023self} implemented a dynamic gallery maintaining a small representative model of people. However, this system lacked long-term robustness. Our approach leverages tracking to ensure long-term control, enhancing the reliability of \reid in industrial environments.

Our method works by combining tracking with \reid. This is a natural progression, as both tasks aim to verify identities across multiple cameras. \cite{leng2019survey} highlighted this integration in their survey, emphasizing the cross-camera identity verification task's importance.
The authors of \cite{suzuki2023runner} propose a method for open-set person re-identification in this direction, using a person detector and a tracker, but their \reid module is not robust to lighting condition changes, as the authors also observed, and very focused on local appearance features specific to their application (\eg the shoes of the runners).

\subsection{Detection and tracking in \reid}
From the point of view of the person detector and tracker, we observe that it is an optimal yet performant solution to use ByteTrack~\cite{zhang2022bytetrack} and YOLOX~\cite{yolox2021}. In fact, other trackers that also perform re-identification during the tracking time have limitations. For instance, FairMOT, introduced by \cite{zhang2020simple}, combines tracking and \reid but is limited by its bottom-up approach, which is prone to false positives and hallucinations. Our system, \ours, addresses these limitations by offering a modular, real-time, and scalable solution tailored for industrial scenarios.
The concept of maximizing the detection quality before performing \reid was also investigated in \cite{mamedov2023approaches}. The authors proposed a filter module to eliminate unwanted images. However, they do not propose a strategy to translate their approach to an open-set scenario.

Overall, our work advances the state-of-the-art by integrating tracking with \reid in an open-set, multi-camera industrial environment, ensuring scalability and robustness in real-time applications.

Finally, our novel dataset, \ice, further contributes to the community by providing valuable data from a practical industrial setting. While complex and large, the existing datasets lack video recording from real industrial settings. Furthermore, the unique conditions of the facility, with many multiple cameras observing the area with various overlap ratios, multiple occlusions, and the diversity of the points of view, gives a strong addition to the available list of \reid datasets.

\section{Method}\label{sec:method}

\begin{figure}[t]
    \centering
    \includegraphics[width=\linewidth]{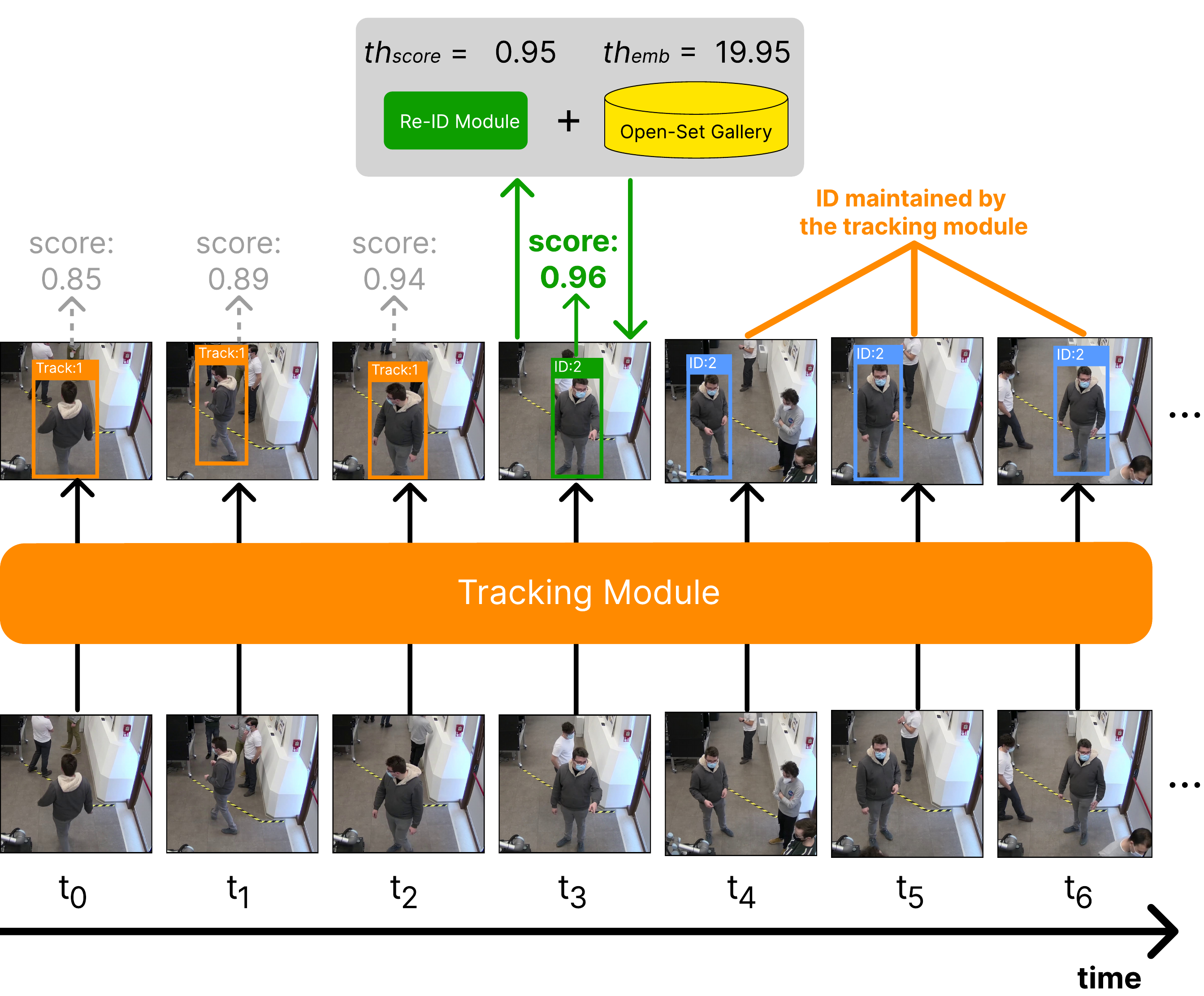}
    \caption{Given a sequence of images, a tracking module extracts the tracks for a person, also providing a confidence score on the detection. If the score is greater than a threshold $th_{score}$, the re-identification module is applied, verifying the person's presence in the Open-Set Gallery. If the person is not present, it is considered a new person. Then, the tracking system maintains the person's identity given by the \reid module.}
    \label{fig:method}
\end{figure}

The person re-identification task consists of accurately identifying and matching individuals across different frames or camera views in a video surveillance scenario.

Our method consists of a pipeline that begins with a tracking module. This module assigns an ID to each detected person in the image, serving as a form of shallow re-identification. However, this ID becomes unreliable when individuals leave and re-enter the camera view, and it is vulnerable to the identity-switch problem, where one track's ID may be erroneously assigned to another. To address this, when a clear image of a person is available, a re-identification module extracts the actual person ID from the current open-set gallery. This ID is then kept in alignment with the track ID, ensuring that the tracker effectively manages both the movement of individuals in the scene and sudden appearance changes (such as variations in lighting conditions, occlusions, etc.).

\subsection{Feature extraction}\label{sec:extractor}

The \reid module extracts and compares appearance features from people's bounding boxes. Given an image $I_i$
 of a person, a model $M_{reid}$ extracts an embedding $e_i$ as follows:
\begin{equation}
e_i = M_{reid}(I_i)    
\end{equation}

with $e_i \in \mathbb{R}^d$, a feature vector of $d$-dimensions representing the appearance of the individual in the image $I_i$.

As module $M$, inspired by~\cite{wieczorek2021unreasonable}, we implement a ResNet50 model~\cite{he2016deep} to learn the identities as aggregated representations in the form of mean centroids.
As each class is represented by a single embedding (class centroid), the retrieval time and storage are significantly reduced, making it less susceptible to a single-image false match.
In this way, the \reid module can learn the appearance of people, specializing in separating in the embedding space the different people (classes). The model training is performed using the Centroid Triplet Loss, following \cite{wieczorek2021unreasonable}.

Real-world images in surveillance settings can have low quality (from the point of view of the people's appearance) for many reasons, such as occlusions or lighting conditions. To make a re-identification system more robust to noise, along with the aggregated centroid representation, we apply the \reid module only to detections with a very high confidence score, namely if the score $s_i > th_{score}$. We consider the confidence score to be a good approximation of high-quality detections in terms of appearance for in-the-wild real-world scenarios. 

Furthermore, to ensure that the \reid module performance is maximized, we propose a solution that relies on a tracking system. 
We leverage the concept of a single-camera tracker as a specialized agent to extract the tracks of people moving in a single camera. Such a system can extrapolate occlusions and use short-term memory appearance features to keep track of a person. The features extracted by a tracking system are not usable as a person re-identifier, as they are shallow re-identifications with short-time memory behaviors. However, a tracking module allows the system to apply the \reid module only on high-quality detections
while leaving it to the tracking system
to keep the assigned ID, taking into account occlusions and variability in terms of appearance (\eg lighting conditions, skewness, \etc) that can lead to \reid performance degradation (see \Cref{fig:method}).

A confidence score threshold $th_{score}$ is defined as a hyperparameter to determine whether the person can be re-identified. 
The rationale is that only if the person is well-visible (\ie, high-quality images with almost no occlusions), it can be surely re-identified. If that is the case, their re-id ID is associated with the track obtained by the tracker. 
\Cref{sec:exps} explores the importance of this threshold in a real-world industrial surveillance scenario.

\subsection{Open-Set Gallery}\label{sec:osg}

After collecting the people tracks and determining which require re-identification, these are provided to the Open-Set Gallery (OSG), an initially empty set of people embeddings.
The OSG compares a probe embedding with the other embeddings previously observed by the system. The feature extractor (see \Cref{sec:extractor}) grants that the probe embedding should be close (using Euclidean distance) to the centroid representing a person in the gallery. 
The OSG keeps track of $K$ embedding of the same person at a time using a circular buffer. Therefore, the OSG has shape $\mathbb{R}^{j\times K \times d}$, with $j$ the number of identities observed up to the current time-step iteration, $d$ the number of the features of the embedding, and $K$ the hyperparameter that represent the history of the person appearance. 
The $j$-th representation of a person is the aggregation of the $K$ embeddings, \ie the centroid.

Given an embedding $e_i$, and a gallery $G \in \mathbb{R}^{j\times K \times d}$, with $j$ the number of current identities and $K$ the buffer length, the retrieval of the ID is represented by the following:

\begin{equation}
    \text{reID}_i = \arg\min_j \left\| e_i - \frac{1}{K} \sum_{k=1}^K G_{j,k} \right\|_2
\end{equation}

If the gallery is empty, $G = \{\}$, or if the minimum distance $\left\| e_i - \frac{1}{K} \sum_{k=1}^K G_{j,k} \right\|$ is higher than $th_{emb}$, then the embedding is considered a new identity and added to the gallery.
The threshold $th_{emb}$ defines the rejection rate of the system. 
The $th_{emb}$ threshold is obtained at the end of the training, computing the average minimum distance between any two classes. 
For our experiments, this threshold was set to 19.95 for the model trained on Market1501~\cite{zheng2015scalable}.

Finally, to consider practical issues in industrial settings, the $K$ embeddings expire after a certain time to prevent the uncontrolled increase of memory usage and to consider the variability of a person's appearance after they have not been seen from the system.
\Cref{fig:method} shows the proposed approach to merge a tracking system with a re-identification module.

\section{Experiments}\label{sec:exps}

We performed two sets of experiments: one investigating the learning of the \reid module to maximize its ability to discriminate the identities, and the other evaluating the performances in the open-set scenario.

\subsection{Learning feature representation}\label{sec:method-training}
To obtain a model that can perform the person re-identification task, we train a model in a supervised way using available closed-set benchmarks.
The first one is the Market1501~\cite{zheng2015scalable} benchmark. This well-known dataset includes 1501 identities and more than 32k bounding boxes.
We also trained on the DukeMTMC-reID~\cite{ristani2016performance} dataset, containing 1852 identities and more than 46k bounding boxes.

The metrics used in the evaluation for the task of \reid are the mean average precision (mAP) and rank-1 accuracy (R1)~\cite{ye2021deep}. See \Cref{tab:closed-set-results} for the results on the closed-set person re-identification of different approaches. 

\begin{table}[t]
    \centering
    \caption{Results of different \reid methods we evaluated on the standard closed-set benchmarks. Note that BPBreID~\cite{somers2023body} was chosen with the lightest backbone to match the computational requirements of the application. Values with (-) were not found in the literature and can't be reproduced due to DukeMTMC~\cite{ristani2016performance} retraction.}
    \begin{tabular}{c|ccc|ccc}
        \hline
        \multirow{2}{*}{\textbf{Method}} & \multicolumn{3}{c|}{\textbf{Market1501~\cite{zheng2015scalable}}} & \multicolumn{3}{c}{\textbf{DukeMTMC~\cite{ristani2016performance}}} \\ \cline{2-7}
        & \textbf{mAP} & \textbf{Rank-1} & \textbf{Rank-5} & \textbf{mAP} & \textbf{Rank-1} & \textbf{Rank-5} \\ \hline
        AGW~\cite{ye2021deep} & 0.878 & 0.951 & 0.973  & 0.796 & 0.890 & - \\
        BPBreID~\cite{somers2023body} & 0.870 & 0.951 & 0.970  & 0.896 & 0.920 & -  \\
        ST-Reid~\cite{wang2019spatial} & 0.955 & 0.980 & 1.00  & 0.927 & 0.945 & -  \\
        CTL~\cite{wieczorek2021unreasonable} & 0.983 & 0.980 & 1.00  & 0.961 & 0.956 & -  \\ \hline
    \end{tabular}
    \label{tab:closed-set-results}
\end{table}

\subsection{Open-Set evaluation}
Regarding the open-set scenario, the available benchmarks for person \reid are collected (and annotated) with the closed-set problem in mind. Hence, they do not depict some of the challenges inherently existing in real-world open-set scenarios~\cite{zahra2023person}. 
While the \reid datasets are oriented to real-world scenarios, they are structured as collections of single images divided by identity, already cropped in single bounding boxes with no temporal continuity. Similarly, datasets used for tracking (like MOT17~\cite{milan2016mot16}), while providing good temporal sequences and some pseudo-labels for the identities (track IDs), have no guarantee of the identities' uniqueness, and they usually don't have varied points of view for the same person, thus providing little challenge \wrt the \reid task.

To evaluate the open-set scenario, we performed two tests: one in which the system ran without the tracking module and score filtering and one with the full system deployed. 
The first test aims to compare the system with the already available datasets in the case they are used as open-set scenarios. The second test runs with the tracking system as if the system was installed in any real-world scenario.

\begin{figure}[t]
    \centering
    \includegraphics[width=\linewidth]{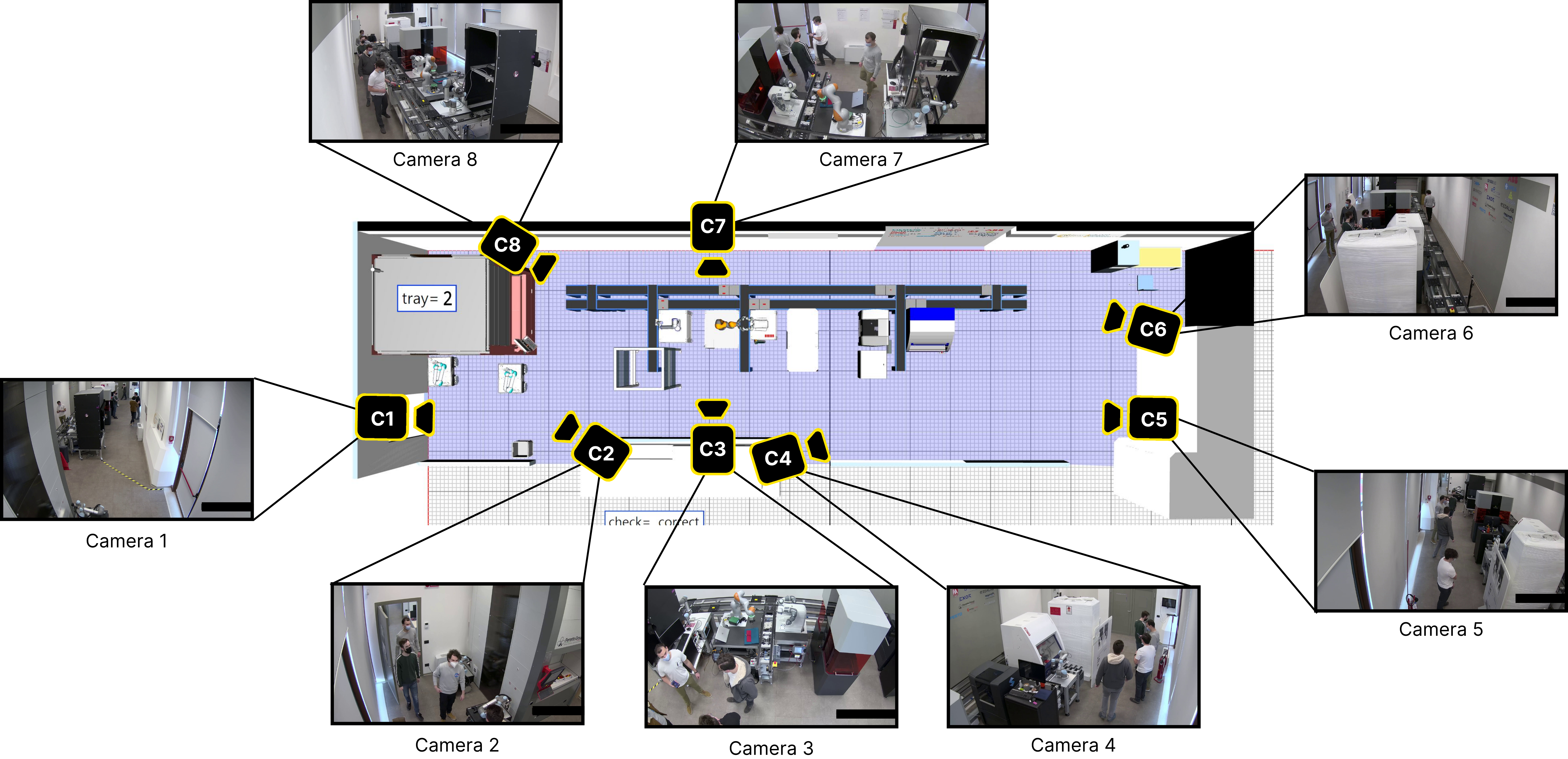}
    \caption{The industrial manufacturing facility's digital twin with the disposition of the multi-camera surveillance system.}
    \label{fig:ice-dataset}
\end{figure}

\begin{figure}[t]
    \centering
    \includegraphics[width=0.9\linewidth]{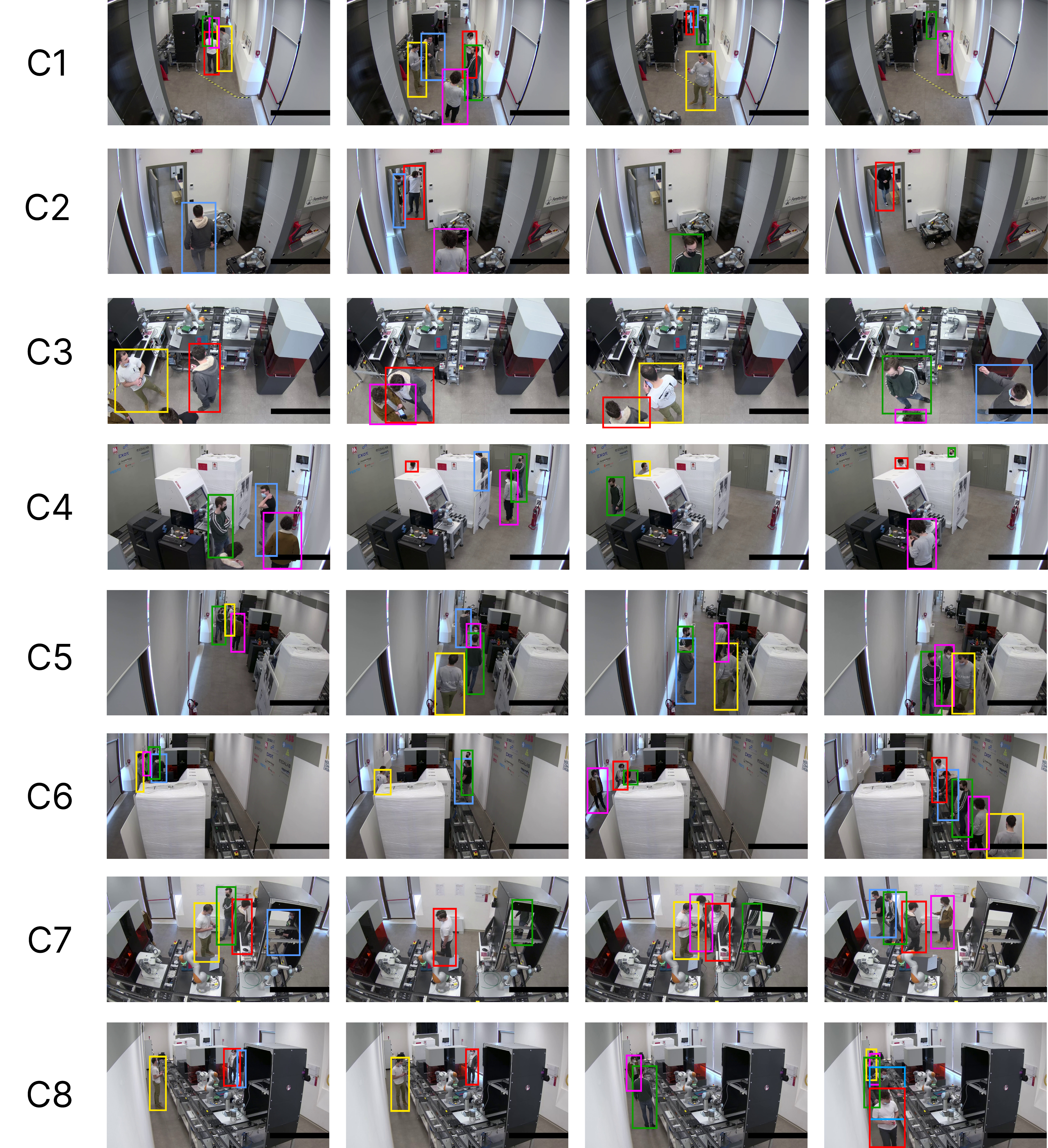}
    \caption{Examples of the images of the \ice dataset. Each line provides examples for a different camera.}
    \label{fig:ice-dataset-examples}
\end{figure}

The first test was performed on the Market1501\cite{zheng2015scalable} dataset by considering it an open-set scenario, meaning the gallery is constructed iteratively. 
Therefore, we removed the tracking component and $th_{score}$ filtering, relying only on the discriminative capacity of the learned features of the \reid module.
In the second experiment, we tested our system on our new benchmark, dubbed \ice, acquired in a real-world industrial manufacturing facility with multiple cameras, occlusions, and non-trivial camera points of view. The dataset consists of $\sim210$k images (18-minute videos) obtained from a multi-camera system with 8 cameras, observing 5 different identities moving freely (alone and in groups) inside the facility. With so many cameras, every person's appearance changes dramatically due to variations of PoV (orientation of the optical axis \wrt ground, alternately long and short views on the environment, \, etc.). The people are also changing clothes (\ie altering their appearance while keeping the same ID) during the visits to make the challenge even more difficult. 
Furthermore, even if there are plenty of occlusions caused both by people moving the room and by the complexity of the environment, the cameras are located in a way such that the entire room is covered by at least one camera.
Examples of the dataset views and camera distribution can be seen in \Cref{fig:ice-dataset} and \Cref{fig:ice-dataset-examples}. 

For evaluation in the open-set scenario, we use TTR (True Target Rate, also called recall) and FTR (False Target Rate)~\cite{liao2014open,leng2019survey}, defined as:
\begin{equation}
    TTR = \frac{N_{t2t}}{N_t}
\quad\quad
    FTR = \frac{N_{nt2t}}{N_{nt}}
\end{equation}

where TTR is the number of accurate verifications. The numbers of probe images from target and non-target people are indicated by $N_t$ and $N_{nt}$, respectively. $N_{t2t}$ denotes the number of accurate verifications that target probe images are matched in the gallery. Similarly, $N_{nt2t}$ denotes the number of false verifications that non-target probe images are treated as the target person.

TTR values can measure performance by verifying target and non-target people and are independent of one-to-one identity correspondence (a closed-set hypothesis). 
According to \cite{leng2019survey}, TTR values (with certain FTR values) are preferred over traditional CMC rates. 
Results on the Market1501 dataset are provided in \Cref{tab:open-set-results}, using a fixed FTR of 1\%. 
Note that the experiments on \ice should be considered zero-shot, as none of the \ice data has been provided in training.

For the person detection and tracking, we used a YOLOX~\cite{yolox2021} combined with the tracker ByteTrack~\cite{zhang2022bytetrack}. The YOLOX detector has been trained on COCO2017~\cite{lin2014microsoft}, achieving an mAP of 51.5, while the ByteTrack tracker has been trained on MOT17~\cite{milan2016mot16}, achieving a MOTA of 80.3. We opted for these models for their reliability, ease of re-training if customization is required, and low inference time (high FPS).

\begin{table}[t]
    \centering
    \caption{Results in the open-set scenario. The results are provided with the closed-set Market1501, extending it as open-set following~\cite{leng2019survey}, and the \ice dataset.}
    \begin{tabular}{c|c|c|c}
         \textbf{Dataset} & \textbf{Precision} $\uparrow$ & \textbf{Accuracy} $\uparrow$ & \textbf{TTR} $\uparrow$ \\ \hline
        Market1501~\cite{zheng2015scalable} & 0.888 & 0.720 & 0.740 \\

        \ice & 0.908 & 0.790 & 0.909

    \end{tabular}
    \label{tab:open-set-results}
\end{table}

To evaluate the impact of our threshold, we perform an exploratory study and present the results in \Cref{tab:abl-th}. We show the impact of this threshold, regulating the image-quality filtering as explained in \Cref{sec:method}. 
The results refer to the \ice dataset, treated as a zero-shot test-only dataset.

\begin{table}[t]
    \centering
    \caption{Evaluation of the impact of the threshold $th_{score}$, \ie on the filtering based on the image quality, on the \ice dataset.}
    \begin{tabular}{c|c|c|c}
         $th_{score}$ & \textbf{Precision} $\uparrow$ & \textbf{Accuracy} $\uparrow$ & \textbf{TTR} $\uparrow$ \\ \hline
         0.99 & \textbf{1.00} & 0.140 & 0.152 \\
         0.98 & \textbf{1.00} & 0.150 & 0.172 \\
         0.97 & \textbf{1.00} & 0.200 & 0.199 \\
         0.96 & \textbf{1.00} & 0.352 & 0.360 \\
         0.95 & \textbf{1.00} & 0.461 & 0.472 \\ 
         0.94 & 0.998 & 0.720 & 0.732  \\
         0.93 & 0.981 & 0.928 & 0.944  \\
         0.92 & 0.996 & 0.949 & 0.950  \\
         0.91 & 0.996 & \textbf{0.964} & 0.965   \\
         0.90 & 0.759 & 0.748 & \textbf{0.981}   
    \end{tabular}
    \label{tab:abl-th}
\end{table}

\section{Implementation details}\label{sec:impl}
The system has been implemented in PyTorch and trained on an NVIDIA RTX 3090. The detector and tracker run at 30 FPS.
The \reid module has an average inference time of 7.28 ms (137 FPS). The system can be further optimized to run on even more constrained devices, like edge devices with GPU capabilities (\eg NVIDIA Jetson), using quantization, pruning, and other techniques~\cite{tinyml}. For instance, using TensorRT, we obtained 20 FPS for the whole pipeline on an NVIDIA Jetson AGX Xavier. This constrained edge device has limited computational resources compared to high-end GPUs, yet still achieves real-time performance for our re-identification system.
The system has been deployed inside the facility using Kubernetes and the docker technology.

\section{Discussion}\label{sec:discussion}

For the closed-set experiments in \Cref{tab:closed-set-results}, we observe that the CTL~\cite{wieczorek2021unreasonable} model performs better than other existing approaches. Therefore, we opted for CTL as the base model for the re-identification task since it appears to learn a better latent representation for discriminating people's identities. 
However, when applied to the open-set modality, it is evident that the performances decrease significantly; 
we advocate this is due to the additional challenge of the open-set scenario.

When running the experiments with the complete \ours, we notice that the scoring filtering is crucial to achieving balanced results. In fact, as the threshold $th_{score}$ decreases, the TTR generally increases. This is consistent with the idea that lower thresholds force the system to use the \reid module more frequently instead of trusting the tracking module to keep the ID previously assigned. This allows the system to capture more true positives, though sometimes at the cost of including more false positives.
Note that when the $th_{score}$ is higher than 0.95, the system is very conservative, reducing even more the accuracy and TTR values, leading to a strong reliance on the tracker rather than a balance between the \reid and tracker module. The accuracy stays at a very high level, meaning that every detection analyzed with the \reid is correctly assigned to the right identity. However, the frequency of the bounding boxes with a high score is limited, especially in complex scenarios.
In fact, we also observed that in the case of severe occlusions, a very high threshold prevents the re-identification, leading effectively to a missed detection (false negative).

The scoring filtering reduces the re-identification of the ``bad'' detections that can possibly lead to errors. However, being more conservative and keeping the threshold to a high value can introduce a delay and potentially a deadlock: we observe that sometimes the detections are below the threshold and, therefore, never re-identified, causing a drop in accuracy.

Nonetheless, when using a balanced value (for instance, 0.91), we notice a very good synergy between the tracking module and the \reid module, obtaining the best results.

We developed the system such that it can run independently on every camera, making it horizontally scalable. In fact, adding a camera to the system requires only replicating the tracking and \reid modules, meaning that \ours can work with any number of cameras and overlapping ratio.

\section{Limitations}\label{sec:limitations}
The system's $th_{score}$ can introduce a delay, which may be higher if the threshold is high. 
Also, considering the \reid module, the maximum number of identities used during the training stage (\Cref{sec:method-training}) can affect the re-identification: if the system is applied to scenarios with a larger number of expected identities \wrt the training set, the risk of identity clashing is higher.
Finally, the thresholds depend on the specific scenario in which the system is deployed, so they must be re-adjusted depending on the scenario.

\section{Conclusions}\label{sec:conclusions}
In this work, we present \ours, a method to perform robust open-set person re-identification in multi-camera scenarios. The system is designed to be easily integrated with industrial environments, scalable, and provides hyperparameters to adapt to various industrial needs. We demonstrate that the system is able to achieve interesting results on available closet-set benchmarks, and the reliability of person detector and tracking systems gives robustness to people's appearance degradation, such as occlusions and lighting conditions, \etc, which are common in industrial environments.
Additionally, we present an industrial \reid and tracking dataset, dubbed \ice, to be used as an open-set industrial dataset for zero-shot open-set \reid validation.

Future work will investigate how to better handle overlapping cameras in the environment. Improving the tracking system such that runs on overlapping cameras rather than single cameras can have a significant impact. 
Furthermore, improvements from the computational point of view, using optimization techniques such as split computing~\cite{cunico2022split,capogrosso2023split} or edge computing~\cite{capogrosso2024machine}, can be beneficial for large industrial applications.
Finally, further tests on the \ice environment are planned, such as, for instance, reducing the number of cameras (that can be done by leaving out observations of specific cameras) or expanding the dataset with more people.

%
%
\bibliographystyle{splncs04}
\bibliography{main}
\end{document}